\documentclass[runningheads]{llncs}
\usepackage[T1]{fontenc}
\usepackage{graphicx,verbatim}
\usepackage{amsmath,amssymb}
\usepackage{booktabs}
\usepackage{multirow}
\usepackage{xcolor}
\usepackage{hyperref}
\begin{document}
\title{From Cross-Section to Cross-Omics:
Conditional Cell Nucleus Image Editing
in Spatial Transcriptomics}
\titlerunning{Cross-Section to Cross-Omics Image Editing in ST}
%

\author{Yuechen Yang\inst{1} \and
Junchao Zhu\inst{1} \and
Junlin Guo\inst{1} \and
Yanfan Zhu\inst{1} \and
Ruining Deng\inst{2} \and
Zhengyi Lu\inst{1} \and
Tianyuan Yao\inst{1} \and
Chongyu Qu\inst{1} \and
Marilyn Lionts\inst{1} \and
Xingyi Guo\inst{3} \and
Yaohong Wang\inst{4} \and
Yu Wang\inst{3} \and
Shilin Zhao\inst{3} \and
Haichun Yang\inst{3} \and
Yuankai Huo\inst{1}}
\authorrunning{Y. Yang et al.}
%
\institute{Vanderbilt University, Nashville TN 37235, USA \and
Weill Cornell Medicine, New York, NY 10065, USA \and
Vanderbilt University Medical Center, Nashville TN 37232, USA \and
The University of Texas MD Anderson Cancer Center, Houston TX 77030, USA\\
}

\maketitle              
\begin{abstract}
Spatial transcriptomics (ST) platforms simultaneously measure gene expression and capture
cell nucleus images at single-cell resolution.
A fundamental property of tissue preparation introduces a confounding relationship between
the two measurement modalities: sectioning a three-dimensional tissue into a thin slice
produces a two-dimensional cross-section whose area depends on the cutting angle.
Larger cross-sections expose more cytoplasm, yielding higher mRNA counts, while
pathomics features such as nuclear area are measured directly from the same 2D boundary.
This sectioning geometry creates an apparent correlation between gene expression and
pathomics that reflects the measurement process rather than independent biology.
Existing gene-to-image methods generate cell images from gene features but do not support
condition-driven editing of existing images across both omics layers.
We apply the Manifold-Aware Diffusion (MAD) framework to this dual-omics
editing setting, enabling targeted modification of one feature group while preserving the
other.
Evaluated on podocyte cells from CosMx SMI kidney tissue, MAD achieves
MAE\,=\,0.6636 and FID\,=\,25.19 for pathomics editing, outperforming all compared baselines.
A controlled simulation study further confirms that the VAE component in MAD better
preserves cross-group independence than CCDM (MAD without VAE).


\end{abstract}

\section{Introduction}
\label{sec:intro}

\begin{figure}[t]
    \centering
    \includegraphics[width=\textwidth]{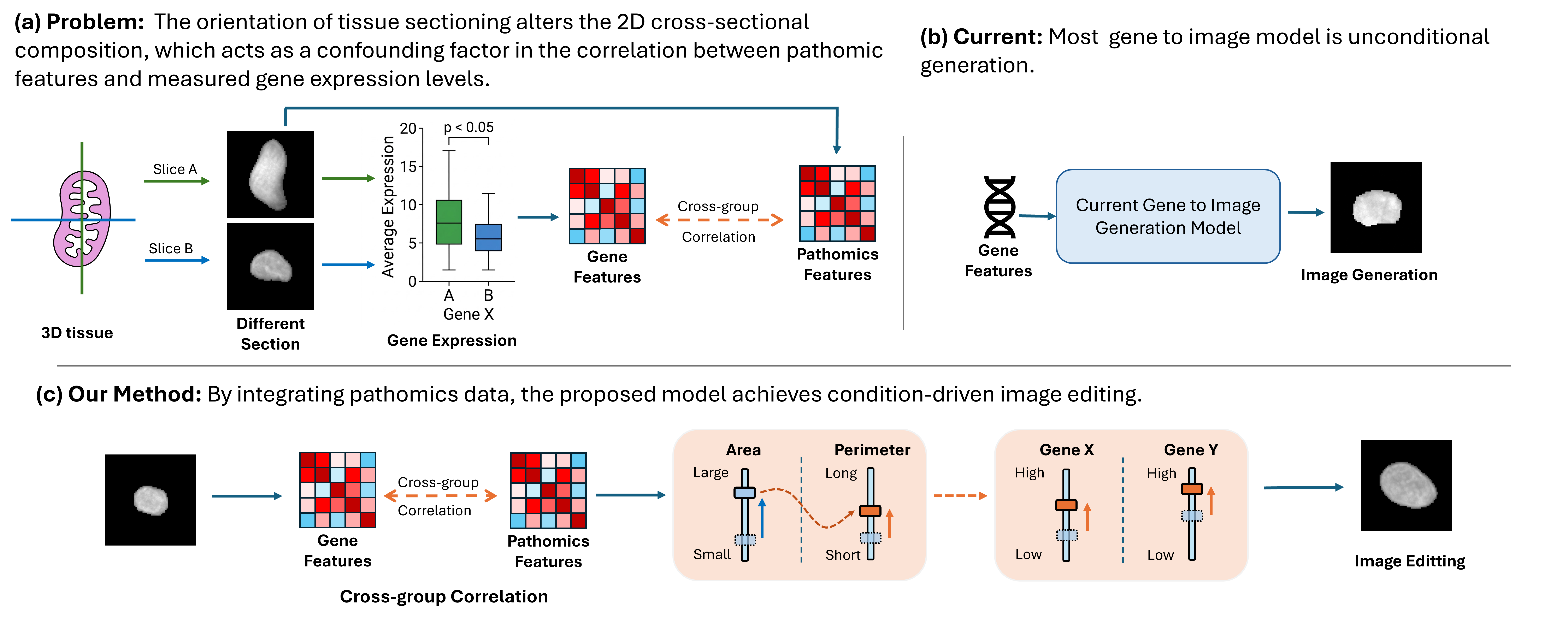}
    \caption{(a) Tissue sectioning mechanism: a 3D cell is cut at varying angles,
    producing 2D cross-sections of different areas. Larger cross-sections yield
    higher mRNA counts and larger nuclear area measurements, creating a
    sectioning-induced correlation between gene expression and pathomics.
    (b) Existing gene-to-image models are generation-only and condition on a single
    omics modality; no prior method performs condition-driven editing jointly across
    gene expression and pathomics.
    (c) Our approach applies MAD~\cite{mad2026} to dual-omics conditional editing:
    given an existing cell nucleus image, one feature group is edited to a target value
    while the other is preserved.}
    \label{fig:intro}
\end{figure}

Spatial transcriptomics (ST) technologies profile gene expression at single-cell
resolution while simultaneously preserving spatial tissue architecture through
fluorescence imaging~\cite{moses2022museum,williams2022introduction}.
Platforms such as the CosMx Spatial Molecular Imager (SMI)~\cite{he2022cosmx}
quantify transcript abundance within individual cells alongside multi-channel
fluorescence images of cell nuclei.
From these images, quantitative morphological pathomic features can be extracted.
Both gene expression and pathomics are therefore natively available per cell,
making them natural conditioning signals for cell image analysis.

A fundamental but often overlooked property of ST data arises from how tissue is prepared
(Fig.~\ref{fig:intro}a).
Sectioning a three-dimensional tissue into a thin slice produces a two-dimensional
cross-section of each cell, and the geometry of this cross-section depends on the
angle and position of the cut.
Cells that are cut near their equator yield larger cross-sections with more exposed
cytoplasm and proportionally higher mRNA capture; cells cut near a pole yield smaller,
gene-sparse cross-sections.
At the same time, pathomics features such as nuclear area and perimeter are measured
directly from these 2D boundaries.
Consequently, both modalities are governed by the same sectioning geometry: larger
cross-sections produce higher apparent gene expression and larger measured
pathomics values.
This creates an apparent cross-omics correlation that reflects the measurement process
rather than an underlying biological relationship.

To our knowledge, no existing method performs condition-driven editing of cell nucleus
images jointly across gene expression and pathomics (Fig.~\ref{fig:intro}b).
Gene-conditioned image generation methods such as GeneFlow~\cite{geneflow2024} and
Genes2\-Morphology~\cite{genes2morphology2025} synthesize cell images from gene features
in a purely generative mode: they produce images from scratch but do not modify an
existing cell image according to a specified target.
Diffusion-based editing models~\cite{rombach2022latentdiffusion} support
condition-driven image modification but are designed for single-modality settings and do
not explicitly account for the independence structure between two distinct omics groups.

In this paper, we formulate the dual-omics conditional editing problem:
given a cell nucleus image and its associated gene expression and pathomics features,
produce an edited image that reflects a target value in one feature group while leaving
the other group unchanged (Fig.~\ref{fig:intro}c).

Our contributions are as follows:
\begin{itemize}
    \item We apply the MAD framework~\cite{mad2026} to dual-omics conditional image
    editing, enabling independent control of gene expression and pathomics features.
    \item We validate on real podocyte data and demonstrate through a controlled
    simulation that the VAE component in MAD better preserves cross-group independence
    than CCDM (MAD without VAE).
    \item MAD achieves MAE\,=\,0.6636 and FID\,=\,25.19, the best performance among all
    compared methods on pathomics editing.
\end{itemize}

\section{Method}

\begin{figure}[t]
    \centering
    \includegraphics[width=\textwidth]{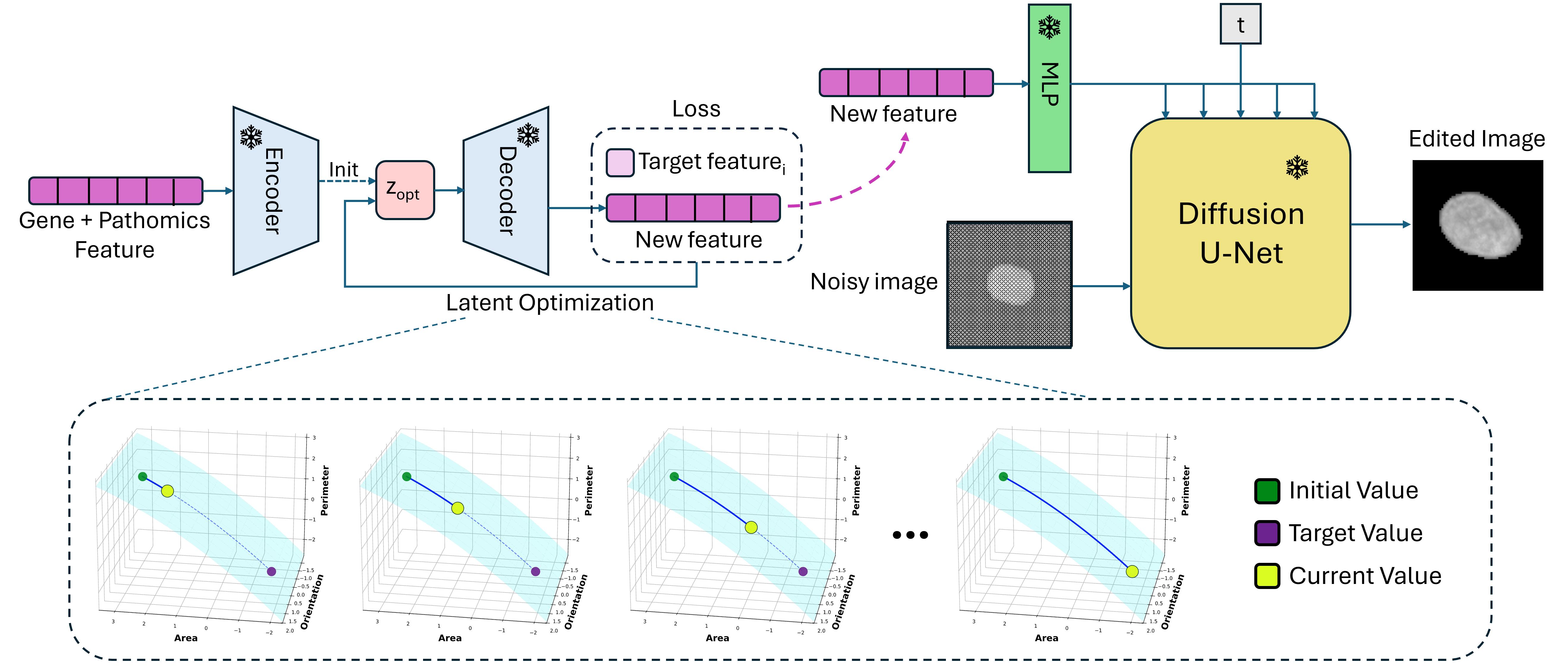}
    \caption{Overview of the dual-omics conditional image editing framework.
    Gene expression $\mathbf{g}$ and pathomics $\mathbf{p}$ are concatenated into a
    joint feature vector $\mathbf{c}$.
    The MAD framework~\cite{mad2026} trains a conditional diffusion model on $\mathbf{c}$
    and a feature manifold VAE that encodes within-group correlations.
    At editing time, a target vector $\mathbf{c}^*$ is projected through the VAE to obtain
    a manifold-consistent conditioning signal, which is passed to the diffusion model to
    generate the edited cell nucleus image.}
    \label{fig:method}
\end{figure}

\subsection{MAD Framework}

We apply the Manifold-Aware Diffusion (MAD) framework~\cite{mad2026} to the dual-omics
editing task (Fig.~\ref{fig:method}).
MAD comprises a conditional diffusion model and a feature manifold VAE.

The conditional diffusion model is a UNet trained to generate
cell nucleus images given $\mathbf{c}$~\cite{ho2020ddpm}.
Following CCDM~\cite{ding2023ccgm}, a mapping network converts $\mathbf{c}$ into a
conditioning embedding that modulates the UNet; classifier-free guidance~\cite{ho2022classifierfree}
is applied at inference.

The feature manifold VAE learns the joint distribution of $(\mathbf{g},
\mathbf{p})$ from observed feature pairs.
At editing time, a target vector $\mathbf{c}^* = [\mathbf{g}_\text{orig}; \mathbf{p}^*]$
(for pathomics editing) is constructed and projected onto the learned feature manifold
by optimizing in the VAE latent space.
The resulting manifold-consistent conditioning vector $\hat{\mathbf{c}}$ is then used
to condition the diffusion model to generate the edited image.
This projection ensures that within-group correlations in the unchanged group are
respected and prevents the editing of one group from inducing spurious changes in
the other---a critical property given the empirical independence between the two groups.

\section{Experiments}

\subsection{Dataset}

We use kidney tissue data from the CosMx Spatial Molecular Imager
platform~\cite{he2022cosmx}, comprising 16 tissue sections (8 diabetic nephropathy,
8 normal) from two samples.
Podocytes are identified by positive WT1 marker expression (WT1\,>\,0).
A tissue-level split assigns 14 sections to training and 2 sections (one diabetic,
one normal) to testing, preventing tissue-level information leakage.
For each cell, a $128 \times 128$ pixel nucleus-centered patch is extracted from the
DAPI fluorescence channel as a grayscale float32 image.

Gene expression features comprise three genes relevant to podocyte biology:
ACTN4 (alpha-actinin-4, associated with podocyte integrity and diabetic
nephropathy~\cite{kaplan2000actn4,thomas2015diabeticnephropathy}),
PODXL (podocalyxin-like, a marker of podocyte apical surfaces), and
VIM (vimentin, an intermediate filament protein).
Pathomics features are extracted using Pyspatial~\cite{yang2025pyspatial} and
all features are Z-score normalized per feature over the training set.

\begin{figure}[t]
    \centering
    \includegraphics[width=0.6\textwidth]{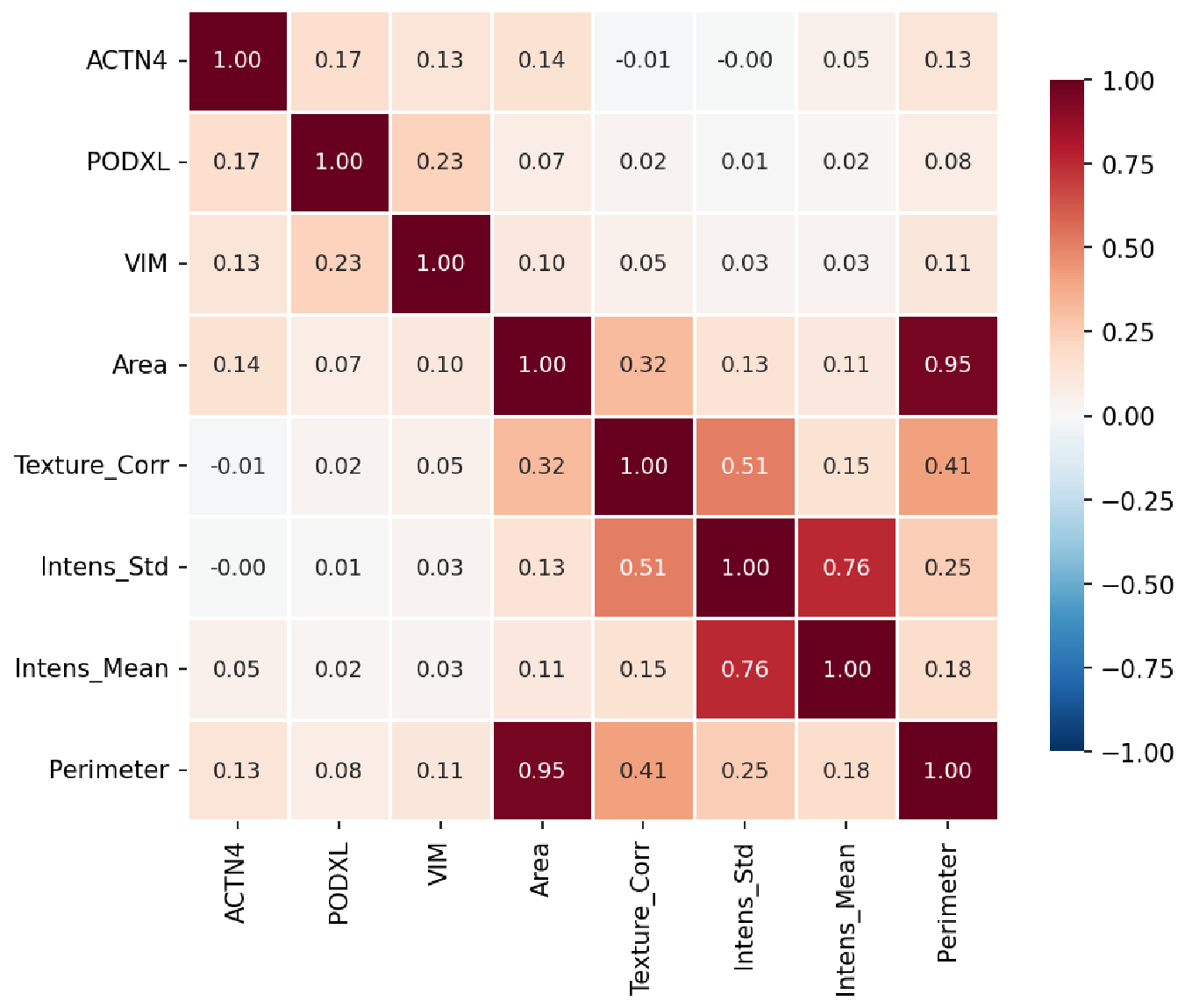}
    \caption{Pairwise Pearson correlation matrix for the 8 features
    (3 genes $+$ 5 pathomics) in the podocyte dataset.
    Strong within-group correlations are observed (Area--Perimeter $r=0.95$;
    IntensityStd--IntensityMean $r=0.76$), while cross-group correlations remain
    near zero ($|r| \leq 0.14$).}
    \label{fig:correlation}
\end{figure}

Figure~\ref{fig:correlation} shows pairwise Pearson correlations between all 8 features.
Within-group correlations are strong, while cross-group correlations remain near zero
($|r| \leq 0.14$).
The general trend that larger area correlates with slightly higher gene expression is
consistent with the tissue sectioning mechanism described in Sec.~\ref{sec:intro}.

\subsection{Baselines}

We compare against four baselines organized into two categories.

\paragraph{Generation-only baselines} (gene expression input; no editing capability):
GeneFlow~\cite{geneflow2024} uses rectified flow with an RNA encoder for
conditional image synthesis from gene features.
Genes2Morphology~\cite{genes2morphology2025} conditions a StyleGAN-T generator
on aligned gene-image representations.
Neither method supports editing of existing images; they are evaluated on image quality
metrics only.

\paragraph{Editing baselines} (joint gene $+$ pathomics conditioning):
Stable Diffusion (SD)~\cite{rombach2022latentdiffusion} is adapted with a
FeatureProjector mapping the 8-dimensional joint feature vector to cross-attention
conditioning; editing uses an image-to-image
diffusion pass on the original cell image.
CCDM~\cite{ding2023ccgm} uses the same two-stage embedding and UNet architecture
as MAD but without the feature manifold VAE, serving as a direct ablation that isolates
the contribution of the VAE component.
MAD~\cite{mad2026} adds the VAE manifold projection to CCDM.

\subsection{Evaluation Metrics}

\paragraph{MAE $\downarrow$} (editing accuracy): Pyspatial is applied to each
generated image to measure its pathomics features; MAE between measured and target values
quantifies how accurately editing achieves the specified target.

\paragraph{FID $\downarrow$} (image quality): Fr\'{e}chet Inception
Distance~\cite{heusel2017fid} between generated and real cell images measures
distribution-level image quality.

\paragraph{SSIM $\uparrow$} (structural fidelity): Structural Similarity Index between
the edited image and the original cell image measures preservation of original cell
structure.

\subsection{Simulation Study}

\begin{figure}[t]
    \centering
    \includegraphics[width=\textwidth]{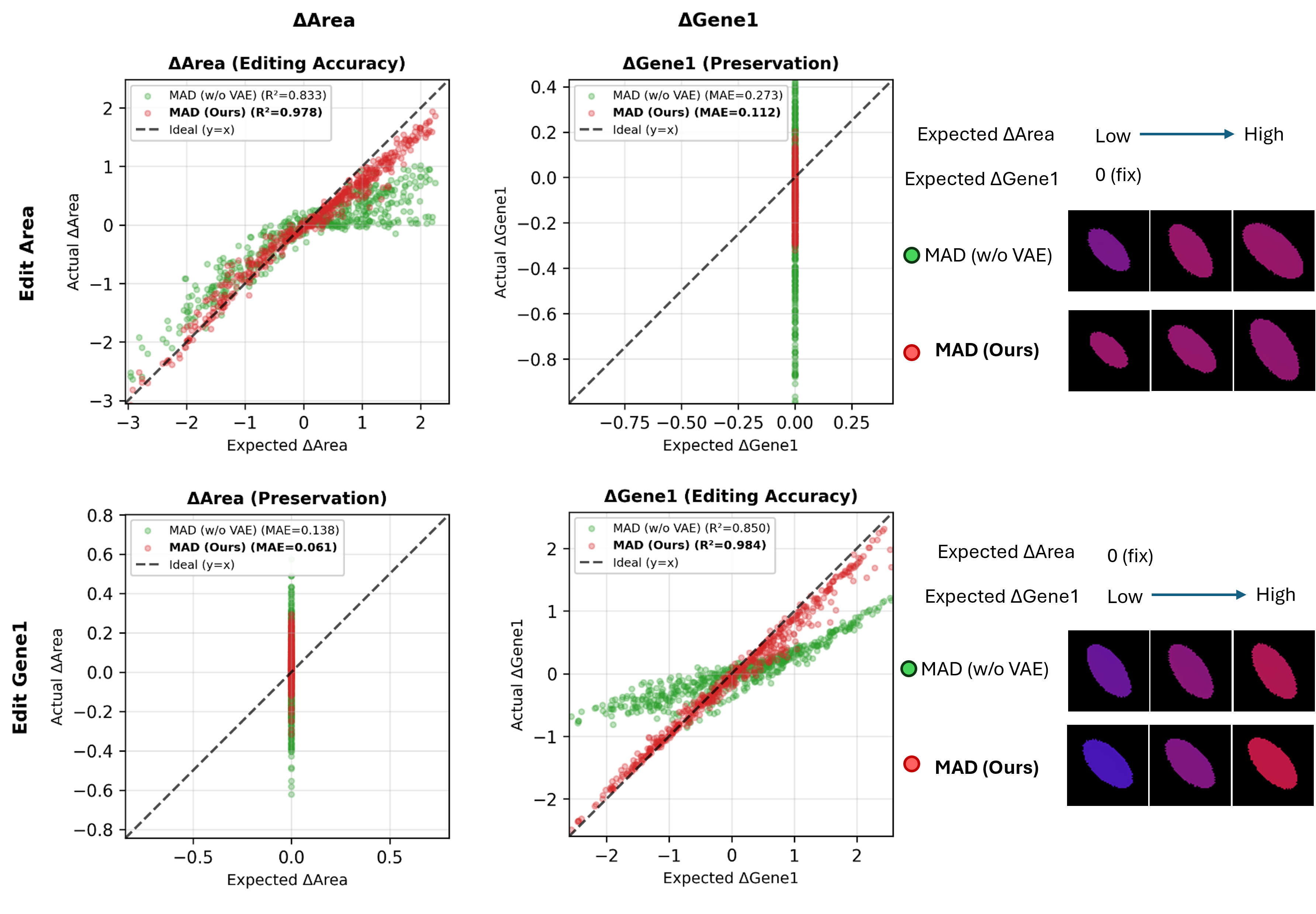}
    \caption{Simulation study: cross-group independence preservation.
    $\Delta$-$\Delta$ scatter plots show the editing-induced change in Group~1 features
    (Area, Gene1).
    Within-group pairs (diagonal blocks) show high correlation; cross-group pairs
    (off-diagonal blocks) should approach zero for a method that preserves independence.
    MAD achieves better cross-group independence preservation
    (cross-group MAE\,=\,0.112 when editing Area) compared to CCDM without VAE
    (cross-group MAE\,=\,0.271).}
    \label{fig:simulation}
\end{figure}

To validate cross-group independence preservation, we construct a synthetic benchmark
with known independence structure.
Four features are organized into two groups:  Area and Gene1; the two groups are statistically independent
by construction, with positive within-group correlation.
Synthetic $128 \times 128$ images are generated from these features via a known process.

For each test sample, Area features are edited to a target value and the induced
change $\Delta_2$ in Gene1 is measured.
An ideal method produces $\Delta_2 \approx 0$ regardless of the editing magnitude $\Delta_1$.
As shown in Fig.~\ref{fig:simulation}:
when editing Area, MAD achieves $R^2 = 0.978$ (editing accuracy) with cross-group
$\Delta\text{Gene1 MAE} = 0.112$, while CCDM achieves $R^2 = 0.831$ with
$\Delta\text{Gene1 MAE} = 0.271$.
When editing Gene1, MAD achieves $R^2 = 0.984$ with $\Delta\text{Area MAE} = 0.061$,
while CCDM achieves $R^2 = 0.850$ with $\Delta\text{Area MAE} = 0.138$.
The VAE component in MAD better preserves cross-group independence compared to CCDM.

\section{Results}

\subsection{Pathomics Editing: Qualitative}

\begin{figure}[t]
    \centering
    \includegraphics[width=\textwidth]{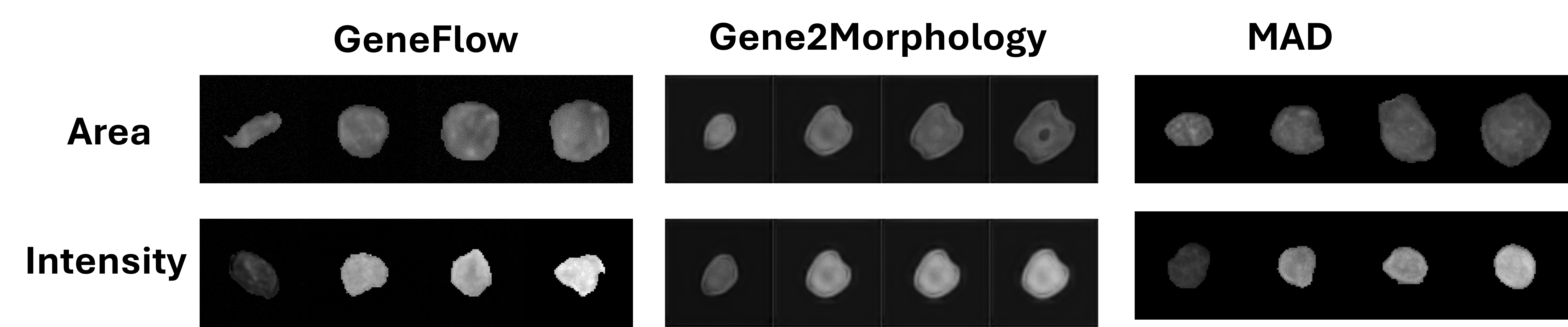}
    \caption{Pathomics editing: generation baselines versus MAD.
    Each row shows a sweep from low (left) to high (right) target values along the
    Area axis (top) and Intensity axis (bottom) for GeneFlow, Genes2Morphology,
    and MAD.
    Generation baselines produce outputs that vary with gene input but do not follow
    the specified pathomics gradient.
    MAD produces progressive morphological change along both editing dimensions.}
    \label{fig:patho_generate}
\end{figure}

\begin{figure}[t]
    \centering
    \includegraphics[width=\textwidth]{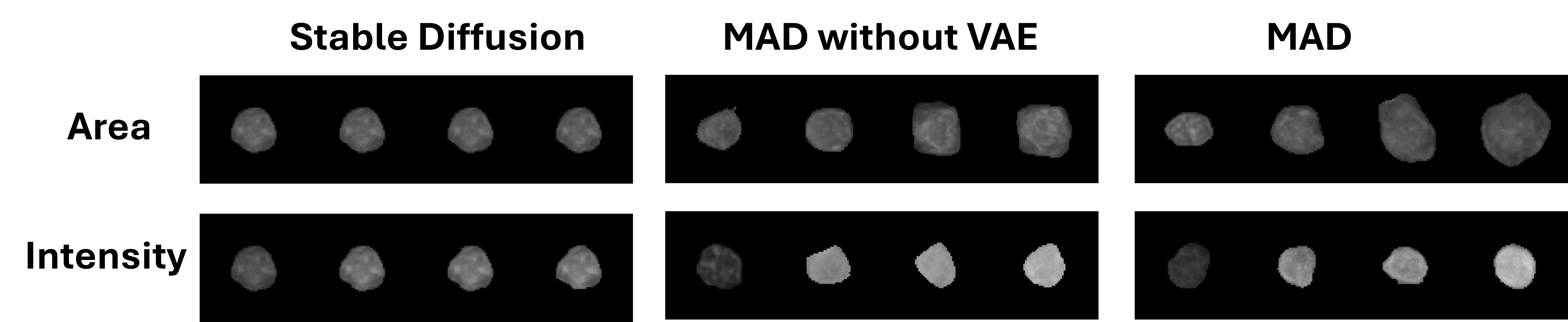}
    \caption{Pathomics editing: editing baselines versus MAD.
    Stable Diffusion (SD) produces minimal visible change across editing steps.
    CCDM (MAD without VAE) shows abrupt intensity variations without smooth structural
    progression.
    MAD achieves smooth, controlled morphological changes along both the Area
    and Intensity editing axes.}
    \label{fig:patho_edit}
\end{figure}

Figure~\ref{fig:patho_generate} compares pathomics editing sweeps for GeneFlow,
Genes2Morphology, and MAD.
As generation-only models, GeneFlow and Genes2Morphology do not condition on pathomics
targets and produce outputs that do not follow the specified gradient.
MAD displays progressive changes in nuclear size and intensity across the editing
range.

Figure~\ref{fig:patho_edit} compares editing models.
Stable Diffusion produces limited structural variation across editing steps.
CCDM shows changes that are not smoothly controlled along the target axis.
MAD achieves gradual, coherent morphological progression for both Area and
Intensity editing.

\subsection{Pathomics Editing: Quantitative}

\begin{table}[t]
    \centering
    \caption{Quantitative pathomics editing results.
    MAE measures editing accuracy (target vs.\ measured pathomics values of generated
    images); FID measures generated image quality; SSIM measures structural fidelity
    to the original cell image.
    $\downarrow$ lower is better; $\uparrow$ higher is better.}
    \label{tab:results}
    \begin{tabular}{llccc}
        \toprule
        Type & Method & MAE $\downarrow$ & FID $\downarrow$ & SSIM $\uparrow$ \\
        \midrule
        \multirow{2}{*}{Generation}
            & GeneFlow~\cite{geneflow2024}
            & 0.8807 & 55.47  & 0.5756 \\
            & Genes2Morphology~\cite{genes2morphology2025}
            & 1.659  & 289.26 & 0.1027 \\
        \midrule
        \multirow{3}{*}{Editing}
            & Stable Diffusion~\cite{rombach2022latentdiffusion}
            & 1.901  & 63.21  & 0.8415 \\
            & CCDM~\cite{ding2023ccgm}
            & 0.9158 & 31.15  & 0.8078 \\
            & \textbf{MAD}~\cite{mad2026}
            & \textbf{0.6636} & \textbf{25.19} & 0.7964 \\
        \bottomrule
    \end{tabular}
\end{table}

Table~\ref{tab:results} reports quantitative results.
Among editing methods, MAD achieves the lowest MAE (0.6636) and FID (25.19), indicating
the highest pathomics editing accuracy and the best generated image quality.
SSIM is slightly lower for MAD than for Stable Diffusion (0.7964 vs.\ 0.8415); Stable
Diffusion preserves more of the original structure but at the cost of poor editing
accuracy (MAE\,=\,1.901).
Generation baselines show high MAE as expected, since they do not condition on pathomics
targets.
CCDM achieves competitive MAE (0.9158) but higher FID (31.15) compared to MAD,
consistent with the simulation findings that CCDM introduces cross-group interference
when the VAE is absent.

\subsection{Gene Editing: Qualitative}

\begin{figure}[t]
    \centering
    \includegraphics[width=\textwidth]{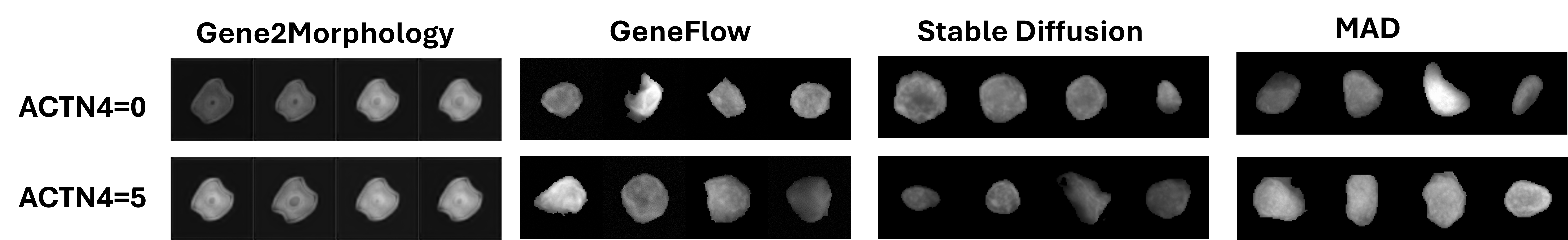}
    \caption{Gene editing results: ACTN4\,=\,0 (normal, left) vs.\
    ACTN4\,=\,5 (disease-associated, right) for Genes2Morphology, GeneFlow,
    Stable Diffusion, and MAD.
    Generation models (Genes2Morphology, GeneFlow) produce outputs without a reference
    image.
    Editing models (SD, MAD) modify an existing cell image.
    MAD produces visually distinct outputs between the two ACTN4 conditions.}
    \label{fig:gene_edit}
\end{figure}

Figure~\ref{fig:gene_edit} shows gene editing results for all models, comparing outputs
for ACTN4\,=\,0 and ACTN4\,=\,5.
Qualitatively, MAD produces visually distinct images between the two ACTN4 conditions;
differences between conditions for Stable Diffusion are less pronounced.

\section{Conclusion}

Tissue sectioning in spatial transcriptomics introduces a cross-omics confound: the
two-dimensional cross-section area co-varies with both gene expression and pathomics
measurements, creating an apparent correlation that is a geometric measurement artifact.
We formulate the dual-omics conditional editing problem and apply the MAD
framework~\cite{mad2026} to enable independent editing of gene expression and pathomics
features in podocyte cell nucleus images from CosMx SMI kidney tissue.
MAD achieves the lowest MAE (0.6636) and FID (25.19) among all compared methods on
pathomics editing.
A controlled simulation study demonstrates that the VAE component in MAD better
preserves cross-group independence compared to CCDM (MAD without VAE).
Future directions include extending the framework to additional omics modalities,
applying it to other cell types and ST platforms, and incorporating spatial neighborhood
features as a third conditioning group for tissue-level analysis.

\begin{credits}
\subsubsection{\ackname} This research was supported by NIH R01DK135597 (Huo), DoD HT9425-23-1-0003 (HCY), NSF 2434229 (Huo) and KPMP Glue Grant. This work was also supported by Vanderbilt Seed Success Grant and Vanderbilt-Liverpool Seed Grant. This research was also supported by NIH grants R01EB033385, R01DK132338. We extend gratitude to NVIDIA for their support by means of the NVIDIA hardware grant.

\end{credits}

%
%
%
\bibliographystyle{splncs04}

\bibliography{midl-samplebibliography}

\end{document}